\documentclass[english]{article}
\pdfoutput=1
\usepackage[preprint,nonatbib]{nips_2018_wider_nonotice}

\usepackage[T1]{fontenc}
\usepackage{color,colortbl}
\usepackage{babel}
\usepackage{verbatim}
\usepackage{url}
\usepackage{amsmath}
\usepackage{amssymb}
\usepackage{graphicx}
\usepackage{setspace}
\usepackage{adjustbox}
\usepackage{overpic}
\usepackage{cancel}
\usepackage{hyperref}
\usepackage{cleveref}
\usepackage{bbm}
\usepackage[utf8]{inputenc}

\usepackage{algorithm}
\usepackage{algpseudocode}

\hypersetup{linkcolor=blue,filecolor=magenta,urlcolor=cyan} 
\usepackage{appendix}
\usepackage{courier}
\usepackage{cprotect}
\usepackage{makecell}
\usepackage{listings}
\usepackage{svg}
\graphicspath{{figures/}}
\usepackage{tablefootnote}

\usepackage[labelfont=bf]{caption}
\usepackage{subcaption}

\usepackage{makecell}
\usepackage{multirow}

\makeatletter
\newcommand{\be}{\begin{eqnarray}}
\newcommand{\ee}{\end{eqnarray}}

\allowdisplaybreaks \numberwithin{equation}{section}
\makeatother

\def\<{\langle}

\usepackage[disable]{todonotes} 
\usepackage{tabularx} 
\usepackage{booktabs} 

\usepackage{amsmath}
\usepackage{empheq}
\usepackage{xcolor}
\definecolor{lightgreen}{HTML}{FFFF99}

\usepackage{geometry}
\usepackage{multirow}
\usepackage{array, makecell}

\title{Harmonic Machine Learning Models are Robust}

\author{
  Nicholas S. ~Kersting\\
  DAP AI Platform \\
  \texttt{nkerstin@visa.com} \\
\And
  Yi ~ Li\\
  DAP AI Platform \\
  \texttt{yili2@visa.com} \\
\And
  Aman ~Mohanty\\
  DAP AI Platform \\
  \texttt{ammohant@visa.com} \\
\And
  Oyindamola ~Obisesan\\
  DAP AI Platform \\
  \texttt{oyobises@visa.com} \\
\And
  Raphael ~Okochu\\
  DAP AI Platform \\
  \texttt{rokochu@visa.com} \\
}

\begin{document}

\maketitle

\section*{Abstract}
\emph{We introduce Harmonic Robustness, a powerful and intuitive method to test the robustness of any machine-learning model either during training or in black-box real-time inference monitoring without ground-truth labels. It is based on functional deviation from the harmonic mean-value property, indicating instability and lack of explainability.  We show implementation examples in low-dimensional trees and feedforward NNs, where the method reliably identifies overfitting, as well as in more complex high-dimensional models such as ResNet-50 and Vision Transformer where it efficiently measures adversarial vulnerability across image classes. }

\section{Motivation and Introduction}

Modern application of Machine Learning (ML) across all industries faces numerous challenges in maintaining quality of live predictions: from the training phase where one must choose the ``best" model within a sea of architectures and hyperparameters to maximize performance without overfitting the training data, while balancing with explainability and fairness in the context of Responsible AI; to the inference phase where, in the face of production latency and throughput constraints, one must efficiently monitor for  performance degradation due to data drift; ideally this latter triggers the model re-training phase, where one must revisit the model with freshly-labelled data, which, however, in many applications such as credit card fraud may not be available till after a significant time lapse, sometimes months later.

To address these challenges, we propose a simple geometric technique which enjoys several mitigating properties:
\begin{itemize}
\item it is model-agnostic (black-box) and unsupervised, i.e., requiring no knowledge of model inner workings, ground-truth labels or other auxiliary data
\item its computation is algorithmically simple, linear in the number of data points tested, and has good statistical sampling convergence
\item it reliably measures relative overfitting between two models on the same training data
\item  it precisely measures model robustness across feature space and can immediately indicate data drift in online monitoring
\item it is indicative of model explainability 
\end{itemize}

\begin{figure}
\centering
\fbox{\rule[-.5cm]{0cm}{0cm} 
\includegraphics[width=0.25\linewidth]{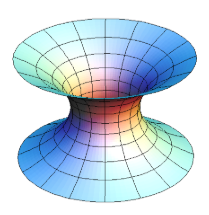}
\includegraphics[width=0.3\linewidth]{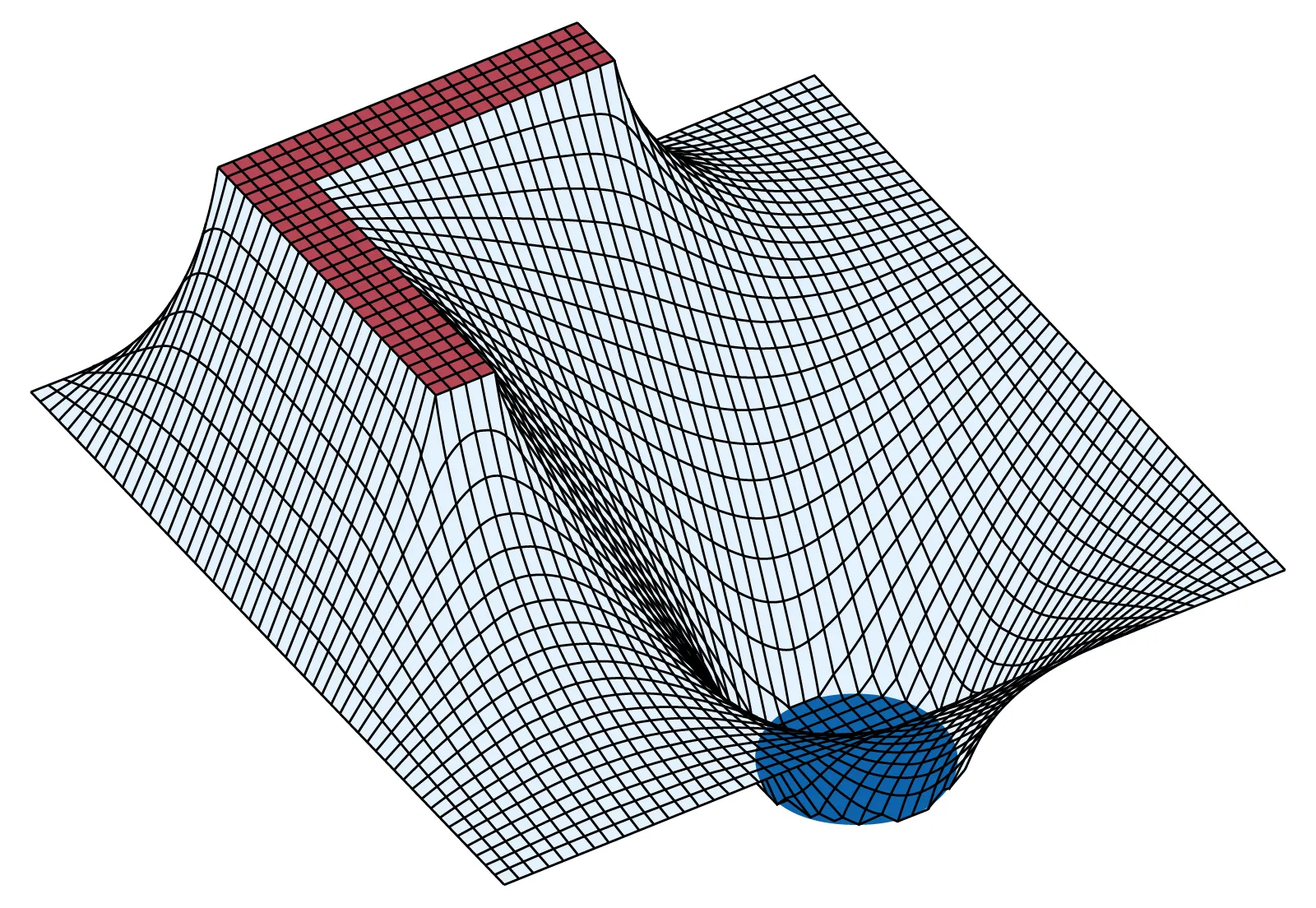}
\includegraphics[width=0.3\linewidth]{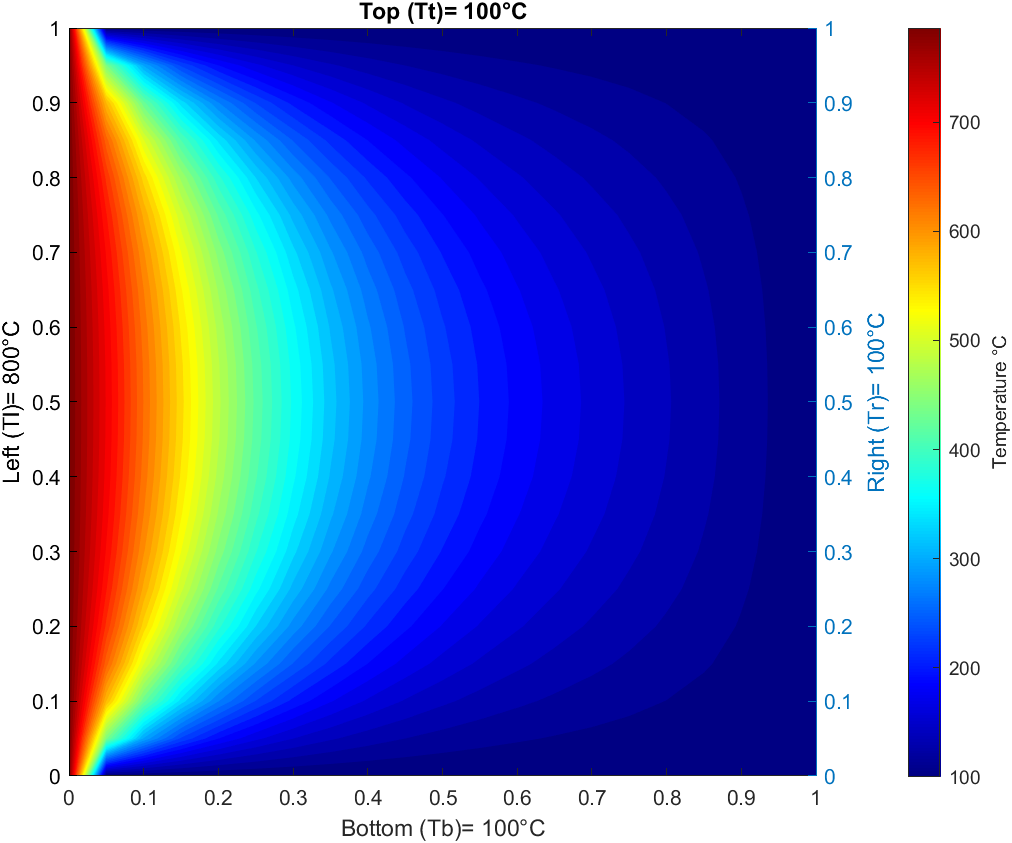}
\rule[-.5cm]{0cm}{0cm}}
  \caption{Examples of harmonic functions that appear in Nature: soap films\cite{harmonic1}, electrostatic potentials\cite{harmonic2}, and heat flows\cite{harmonic3}.}
 \label{fig:har}
\end{figure}

The particular proposal is to measure the ``harmoniticity" of the model, specifically the degree to which the model function $f$ satisfies the harmonic property,
\begin{equation}
\label{harmonic}
\nabla^2 f = 0
\end{equation}
Functions which satisfy (\ref{harmonic}), i.e, ``harmonic functions'', occur frequently in physics as solutions to equilibrium problems involving minimization of energy, e.g. soap bubbles stretched on a boundary, electrostatic field configurations, and heat flow (see Fig~\ref{fig:har}). They form smooth interpolations between boundary values, and most importantly to our present discussion exhibit the ``mean-value property",
\begin{equation}
\label{average}
f(x) = \frac{1}{V r^n} \int_{B(x,r)} f dV
\end{equation}
which, in plain English, says that the value of the function at any point is the average of the function over a ball of any radius $r$ surrounding the point (incidentally, (\ref{harmonic}) and (\ref{average}) are equivalent definitions)\footnote{An easy informal way to see this equivalence is to recognize (\ref{harmonic}) as the divergence of the gradient ($\nabla^2 f = \bigtriangledown \cdot \bigtriangledown f$): the gradient expressing the change of $f$ in all directions, if its divergence is zero then 'change in f' neither flows into or out of any given point, hence the average change of $f$ on any ball around that point is 0, relative to its value at the point.}. The  metric we propose, "anharmoniticity" or $\gamma$ for brevity, measures how well (\ref{average}) is satisfied over feature space, computing the difference between the function and its ball-averaged value:
\begin{equation}
\label{gamma}
\gamma(x,r) \equiv |f(x) - \frac{1}{V r^n} \int_{B(x,r)} f dV|
\end{equation}

As the behavior of $f(x)$ may vary wildly over feature space, so in general will $\gamma(x,r)$ depending on the degree to which $f$ behaves in accordance with (\ref{average}) for some reasonable fixed choice of $r$. By association this will indicate which regions of feature space are ``more harmonic" for this model, and the average value of $\gamma(x,r)$ over feature space provides a summary ``anharmoniticity" metric. 

At this point, we can verify the above claimed mitigating properties of this metric:
\begin{itemize}
\item Measuring $\gamma$ as per (\ref{gamma}) requires nothing more than black-box access to the model, as is expedient in an inference setting; this also allows testing of closed-source models without having to request access to model details, facilitating efficiency of testing and helping to keep the industry honest.
\item As $\gamma$ only requires computing the average value of $f$ at a number of data points approximating a ball, this is linear in the number of points and amenable to sampling.
\item $\gamma$ is proportional to the complexity of the decision surface; in particular for a binary classifier it is proportional to the length of the decision boundary, which is positively correlated with overfitting (see Appendix~\ref{app:boundary}).
\item   if $\gamma$ changes over time in online events, there must be data drift; this is great for online monitoring where it is paramount to raise alerts as soon as a performance issue arises. If production monitoring shows an increase in $\gamma$, one can pinpoint the data points responsible and investigate that region of feature space more fully for counterfactuals --- this might trigger the need for re-training with more data or modeling in that region.
\item harmonic functions are natively explainable since, by the mean value property, the 'explanation' of any point is that it is the average of the points around it, which in turn are `explained' by their neighboring points, etc., all the way up to the feature boundaries which have values fixed by some standard. The premier example is the linear function, of course trivially harmonic by  (\ref{harmonic}) and explainable by direct proportionality. Thus, the closer $\gamma$ is to zero, the more explainable the model will be. Conversely, the more a model fails (\ref{average}) at some point the more difficult it may be to explain, e.g., in the fraud domain if the average of several non-fraudulent events was predicted to be fraudulent.
\end{itemize}

In short, $\gamma$ is a proxy for the measure of model robustness in stability of prediction, resilience to data drift, and ease of explainability.

Note, however, that the aim of this programme is certainly not to have a model be completely harmonic --- indeed  by (\ref{average}) it is easy to check that pure harmonic functions can have no local minima or maxima, and that is too restrictive for real-world models.

Yet, we believe a robust, explainable ML model should be at least locally close to harmonic in \textit{most} of feature space\footnote{This can be made precise by choosing a value of $r$ that is 'small' relative to typical distance between data points. Though as we'll see below, the exact choice is not critical.}, especially for production applications where stability is business-critical. Most  real-world models will of course deviate from pure harmoniticity, but $\gamma$ gives us a way to  track and quantify the deviation.

Finally, as far as we know, this is the first instance in the literature of an overarching, limiting {\em algebraic} standard on a ML function for the purpose of quality and stability. Our choice of the harmonic property is not in any way sacrosanct, but it is afterall intuitive and accessible for quick and direct testing via the Mean Value Property (\ref{average}), giving correlation with stability and ease of interpretation. It may be that some other class of functional algebraic constraints also captures this and more, a topic we leave open to the community to explore.

\section{Related Work}

Techniques to measure goodness of a predictive model are of course as old as the field of Machine Learning itself, traditionally centered on time-tested metrics such as precision, recall, F-score, AUC, etc. where the ground-truth labels of a test set are known. 

On the other hand, we are chiefly concerned with the real-world problem of measuring model robustness without access to ground truth labels, as for example occurs in a purely online inference environment with only black-box access to the model in conjunction with a live data stream. For this is the real, minimal environment in which most practioners and end users of ML operate.
Statistical techniques such as outlier or anomaly-detection \cite{bouman2024unsupervised} and distributional shift \cite{wiles2021fine} enjoy usage here to  give important hints of, but not true indications of, model robustness. Gradient-based methods such as PDP and ICE\cite{goldstein2015peeking} where one looks for sudden changes in the decision function over feature space likewise may provide hints of  robustness changes, though this differs from the current proposal as  $\gamma$ is measuring more than just the local feature sensitivity in the function, which may in fact be proper and desired behavior: rather, $\gamma$ is measuring departure from {\em explainable} sensitivity as one sees in harmonic functions obeying the mean-value property (\ref{average}).

Existing work \cite{bastani2016measuring} as well as a recent survey \cite{guo2023comprehensive} reviews 23 metrics which are useful in the online inference setting, e.g., Average Confidence of True Class (ACTC) and Noise Tolerance Estimation (NTE). Within the black-box setting the metrics are essentially measuring how readily the predicted class label changes across feature space either due to targeted gradient-based search or random perturbation. These fit in the realm of Adversarial Machine Learning \cite{costa2023deep} which has blossomed into its own subfield, quite rightly dedicated to understanding the vulnerability of popular ML models to attacks \cite{szegedy2013intriguing}\cite{goodfellow2014explaining} based on perturbing input data points. Recent work has found  that adversarial weakness becomes more prevalent with increasing number of feature dimensions \cite{wang2016theoretical} \cite{gilmer2018relationship}; unsurprisingly this is most apparent in image classification tasks, the premier testing-ground of adversarial ML, as each data point can easily contain thousands to millions of features (pixels); it is important to note, however, that many other domains, e.g., financial  modeling, can contain thousands or more features  and likewise be highly vulnerable to attacks \cite{goldblum2021adversarial}. Adversarial analyses also typically focus on class label changes, differing  from our metric which is more precisely measuring the numerical stability of the prediction (logit) according to the standard of harmonic geometry, and not merely the crossing of the such predictions over discrete thresholds leading to class label changes. 

There have for a number of years been works focusing on the relationship between stability and geometry of the classifier \cite{caramanis2011robust}\cite{fawzi2016robustness}\cite{moosavi2017robustness}: what these works find is that there is a correlation between adversarial weakness and curvature of the decision surface. This has even fueled investigation into a new way of classification using the average or majority-vote in a hypercube neighborhood of each point \cite{cao2017mitigating}. 
This is corroborated by the present study as well, for harmonic functions describe minimal surfaces with constant mean curvature\cite{talenti1982note}, hence should have minimal adversarial weakness.

\section{Method}

Computing $\gamma$ for a model as per its mathematical definition (\ref{gamma}) is an extremely simple and straightforward procedure, which we detail below in pseudocode:

\begin{algorithm}
\caption{Computation of $\gamma$ at a point x in feature space}
\begin{algorithmic}[1]
\Procedure{$\gamma(x,r)$}{}
\State $\textrm{ballPoints} \gets \mathrm{Ball}(x,r)$
\State $\textrm{N} \gets size(\textrm{ballPoints})$
\State $\textrm{ballValue} \gets 0$
\For{\texttt{each $\mathrm{point}$ in $\mathrm{ballPoints}$}}
\State $\textrm{ballValue} ~+=~  f(\mathrm{point})$
\EndFor
\State $\textrm{ballAvg} \gets \textrm{ballValue}$/N
\State \Return $|f(x) - \textrm{ballAvg}|$
\EndProcedure
\end{algorithmic}
\end{algorithm}

One can then average $\gamma(x)$ over a region of feature space to get $\gamma$ for that region, for example the convex hull of a training set or all feature points seen in some inference production window. 

In the above algorithm, the primary consideration is how to get the ball of radius $r$ around the point x, i.e., $\mathrm{Ball}(x,r)$, remembering that in general x is a vector in some possibly high number of dimensions. For any digital computation we will of course have to approximate a continuous ball with a discrete number of points. The easiest-to-code solution is to construct some large number of random vectors (each normalized to some small magnitude $r$) of the same dimension around x, hoping for isotropy and centrality (zero overall vector sum). The random-walk behavior of N random vectors will however doom one to a $\sqrt{N}$ bias in one direction or another (see Appendix~\ref{app:functions}).

\begin{figure}
\centering
\fbox{\rule[-.5cm]{0cm}{0cm} 
\includegraphics[width=0.7\linewidth]{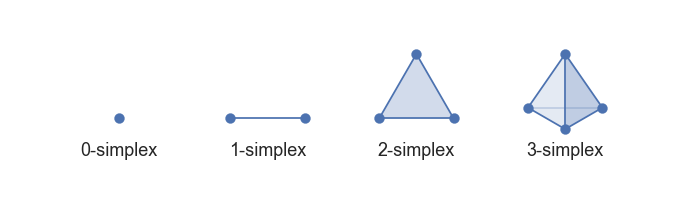}
\rule[-.5cm]{0cm}{0cm}}
  \caption{The first few n-simplices ...}
 \label{fig:simplex}
\end{figure}

 A better solution from a theoretical perspective is to form the "n-simplex" around each point. In two dimensions, for example, the 2-simplex is an equilateral triangle; in three dimensions a tetrahedron, etc. (see Fig. \ref{fig:simplex}). In any arbitrary number of dimensions, the n-simplex centered about a point will be maximally symmetric, hence ideally space-covering. One can further add balanced rotations of the basic n-simplex, the more of which you add the closer the discrete approximation converges to the continuous ball.
 
With a bit of linear algebra  to compute the n-simplices (see Appendix~\ref{app:simplex}), numerical trials indeed show that n-simplices symmetrically cover space much more effectively than n-random-vectors and, when used to approximate the ball in our algorithm, accurately identify pure harmonic versus non-harmonic functions, the details of which the interested reader may refer to in Appendix~\ref{app:functions}. What we would like to focus on in the remainder of this paper is actual ML models, as readers will find this most applicable to their work.

\section{Application to low dimension models}
\label{sec:ml}

Moving right into the application to real ML models, let us apply the foregoing first to basic models targeting a small, well-understood dataset in the ML community: the Wine  dataset, describing 13 features of three different wines grown in the same region of Italy  \cite{mea1991parvus}. We will focus on just two of those dimensions, however, ``flavanoids" and ``OD280/OD315 of diluted wines",  in order to demonstrate a model taking a two-dimensional input vector to a scalar output (the wine class label).

The goal will be to determine whether models trained with these features on this data are well-fit or overfit, hence vulnerable to adversarial attacks or production under-performance, purely from inference on the test-set without referencing ground-truth labels.

\subsection{Gradient Boosted Decision Tree}

\begin{figure}
\centering
\subfloat[]{\includegraphics[width=0.5\textwidth]{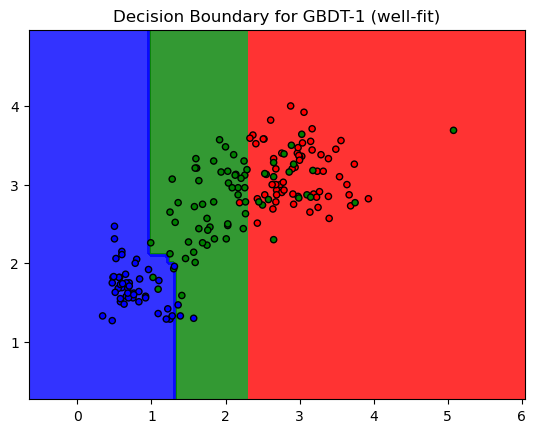}}\hfill
\subfloat[]{\includegraphics[width=0.5\textwidth]{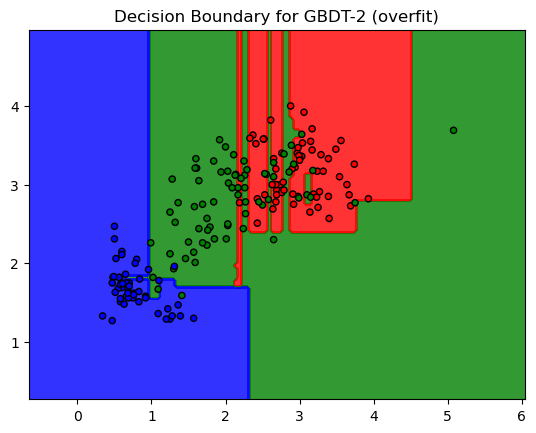}}\\
\subfloat[]{\includegraphics[width=0.5\textwidth]{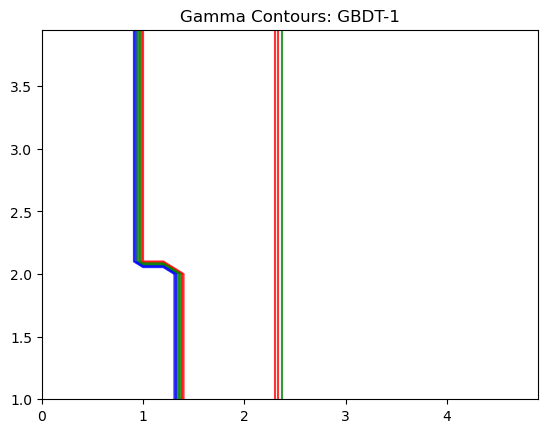}}\hfill
\subfloat[]{\includegraphics[width=0.5\textwidth]{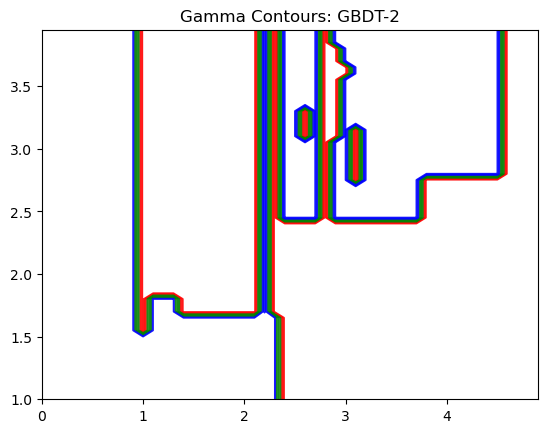}}\hfill
\caption{Decision functions and gamma contours of the two classifiers: GBDT1 (left), and overfit version GBDT2 (right).}
 \label{fig:fits}
\end{figure}

We start out with a well-understood, explainable ML model as well, the Gradient Boosted Decision Tree (GBDT).
After splitting the original data 80/20 to a Train/Test set\footnote{For this toy example with extremely sparse data we opted out of a Validation set for simplicity.}, we train two models: the first, ``GBDT-1",  with hyperparameters optimized on a grid search with 10-fold cross-validation, and the second, ``GBDT-2", chosen with more extreme hyperparameter values, as shown in  Table~\ref{tab:gbdt}.
Intuition should tell us that GBDT-2 will be overfit, and the performance on the Train/Test sets shown bespeak to this, 
attaining a high training accuracy (100\% in fact) while suffering on the test set with only 80\% accuracy. GBDT-1, meanwhile, more reasonably attains a weighted accuracy of 85\% on the train set and 83\% on the test set. With the explicit labels on the test set, the case against the overfit model is obvious. But what if that test set were not available? 
An important hint is in the shapes of the decision boundaries for these two classifiers: note how the shape of the overfit model's decision boundary Fig.~\ref{fig:fits}(b) is much more complicated than that of the well-fit model in Fig.~\ref{fig:fits}(a). What we will show  is that computing $\gamma$ on the relevant region of feature space leverages this fact to identify the overfit model\footnote{The conventional way is to of course use a validation set to determine that one model was overfit; we don't discount that here, but wish to show that gamma provides an additional metric which is moreover applicable to unlabeled data.}.

\begin{table}
\centering
\caption{Model parameters and performance for two different models trained on the Wine dataset.}
\label{tab:gbdt}
\begin{tabular}{|l|r|r|} 
\hline
  & GBDT-1 & GBDT-2 \\ 
\hline
max\_depth  & 1  & 100  \\
n\_estimators  & 5  & 200  \\
min\_samples\_split  & 2  & 2  \\
learning\_rate  & 0.1  & 1  \\
Train accuracy & 85\%  & 100\%  \\
Test accuracy  & 83\%  & 80\%  \\
\hline
$\gamma$ (r=0.05) & 0.014(2)  &  0.051(2) \\
\hline
\end{tabular}
\end{table}

\begin{figure}
\centering
\fbox{\rule[-.5cm]{0cm}{0cm} 
\includegraphics[width=0.5\linewidth]{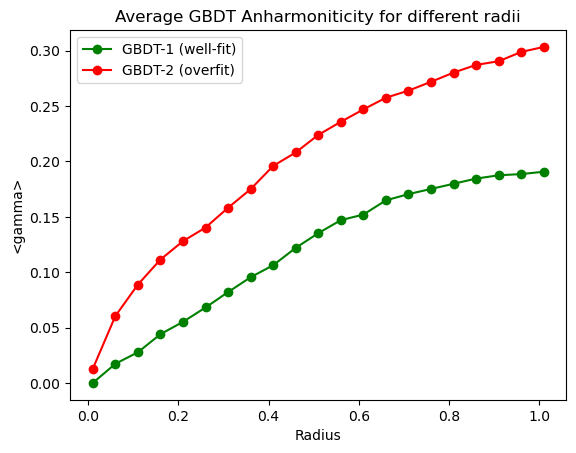}
\includegraphics[width=0.5\linewidth]{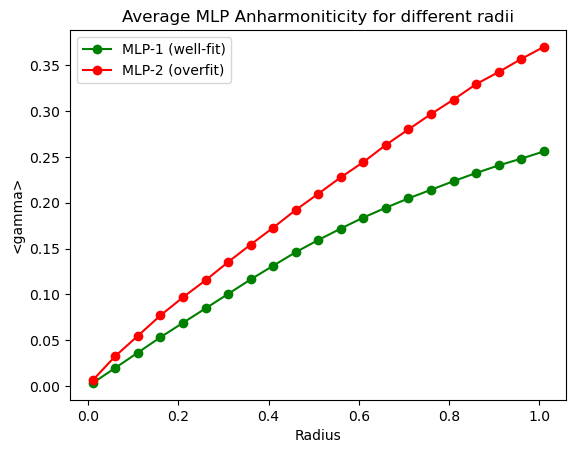}
\rule[-.5cm]{0cm}{0cm}}
  \caption{Average anharmoniticity for the GBDT well-fit and overfit models (left) and MLP models (right) at various choices of the radius parameter $r$. It is significantly greater for the overfit function at any choice of radius $r$ (statistical error is too small to show).}
 \label{fig:radii}
\end{figure}

Our procedure, then, is to compute $\gamma(x)$ on a grid over the boxed region [[0,5],[1,4]] which safely encloses all data points for each classifier's decision function. As this is a 2D problem, the simplices that we form around each point are 2-simplices (equilateral triangles), and we also add the flipped 2-simplices around each point for better 'ball coverage'. Choosing a ball radius of 0.05 (more on this choice shortly), for example, we obtain the $\gamma$ contours of Fig.~\ref{fig:fits}(c)-(d). Notice how $\gamma$ is non-zero only around the decision boundary, which for the overfit function in Fig.~\ref{fig:fits}(d) is much more convoluted. Indeed, the average value of $\gamma$ over this region is nearly 4 times higher for the overfit function as reported in Table~\ref{tab:gbdt}. The choice of radius $r$ has some, but no major effect on this result: at any nonzero value of $r$,  the overfit function has a significantly higher $\gamma$ (see Fig.~\ref{fig:radii}).

\subsection{Multilayer Perceptron}

Next we consider a potentially more powerful, but more opaque model: the feedforward neural network, aka Multilayer Perceptron (MLP). Paralleling our discussion for the GBDT models, we train one reasonable model ``MLP-1" and one other model ``MLP-2" likely to overfit. The parameters of these models is shown in Table~\ref{tab:nn}. By design, MLP-2 is under-regularized and over-parameterized with 3 high-dimensional hidden layers as opposed to MLP-1's 1 hidden layer.
And as before with the GBDT models, the overfit model MLP-2 predictably performs better on the train set but worse on the test set than MLP-1, having a more convoluted decision boundary that also registers a significantly higher $\gamma$ than MLP-1. 
 
\begin{figure}
\centering
\subfloat[]{\includegraphics[width=0.5\textwidth]{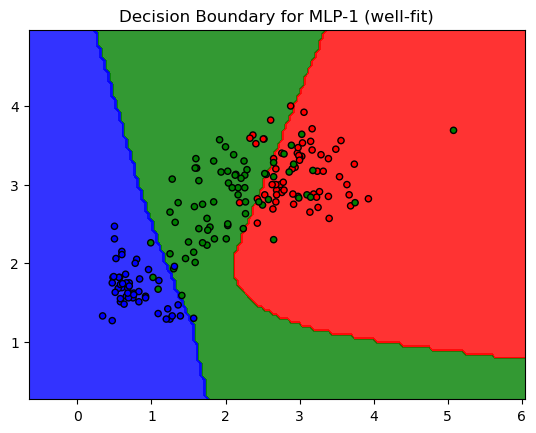}}\hfill
\subfloat[]{\includegraphics[width=0.5\textwidth]{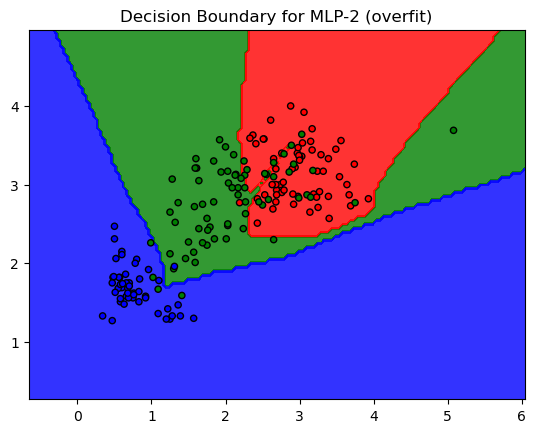}}\\
\subfloat[]{\includegraphics[width=0.5\textwidth]{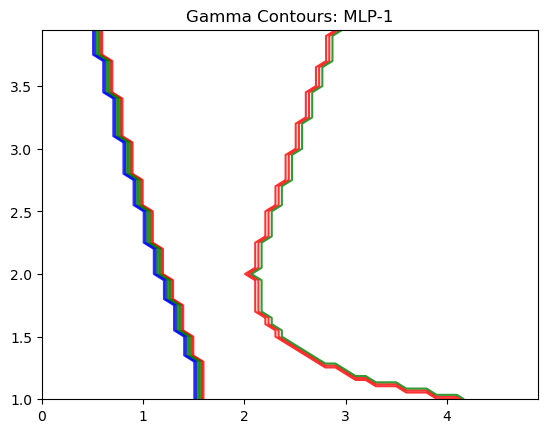}}\hfill
\subfloat[]{\includegraphics[width=0.5\textwidth]{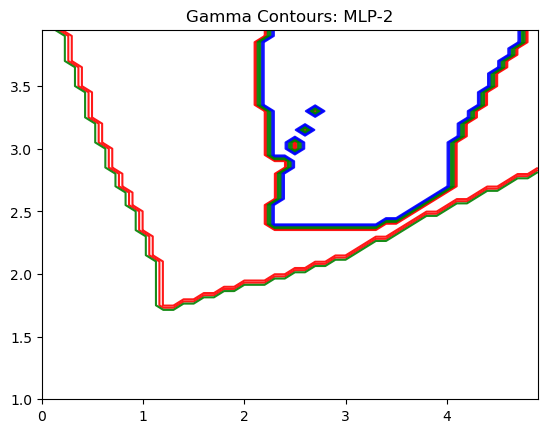}}\hfill
\caption{Decision functions and gamma contours of the two classifiers: NN-1 (left), and overfit version NN-2 (right).}
 \label{fig:nfits}
\end{figure}

What's interesting is that both GBDT-1 and MLP-1 have the same Test performance of 83\%, but GBDT-1 has a slightly better $\gamma$ of 0.014 versus MLP-1's $\gamma$ of 0.016. One can easily see this difference from the lengths of the decision boundaries: the GBDT boundary in Fig.~\ref{fig:fits}(c) consists of nearly straight lines while the MLP boundary in Fig.~\ref{fig:nfits}(c) is more curved. This illustrates how a model trainer might, from the perspective of robustness, prefer the GBDT model for this dataset.

\begin{table}
\centering
\caption{Model parameters and performance for two different MLP models trained on the Wine dataset.}
\label{tab:nn}
\begin{tabular}{|l|r|r|} 
\hline
  & MLP-1 & MLP-2 \\ 
\hline
max\_iter  & 200  & 1000  \\
layer dims  & (2,100,1)  & (2,100,500,1000,1)  \\
initial learning\_rate  & 0.001  & 0.01  \\
optimizer & adam & adam \\
alpha & $1 \cdot 10^{-4}$ & 0 \\
validation\_fraction & 0.1 & 0 \\
Train accuracy & 82\%  & 86\%  \\
Test accuracy  & 83\%  & 79\%  \\
\hline
$\gamma$ (r=0.05) & 0.016(1)  &  0.027(1) \\
\hline
\end{tabular}
\end{table}

It should be clear from the above, then, that the Harmonic Robustness metric clearly works on simple ML functions in low numbers of dimensions, where we can visually confirm areas of feature space which are more robust and compare robustness of different models over the same space.

\section{Application to high dimension models}

As we have seen, the application of  Harmonic Robustness to models with a low number of input and output (in fact, scalar) dimensions is direct and intuitive. In this section we will consider the more complex case of higher-dimensional models to illustrate how the technique adapts.

The challenge of models over a larger number of dimensions (into the thousands or even millions) is three-fold: 
\begin{enumerate}
\item higher-dimensional simplices are more expensive in compute and storage
\item high-dimensional models are typically more complex  and take longer to run
\item $\gamma$ itself might be high-dimensional and interpretation is not straight-forward
\end{enumerate}
For the first challenge, Mathematics is actually kind to us, where it turns out that for large n the n-simplex is approximately the same as the vertices of the n-dimensional hypercube (see Appendix~\ref{app:simplex}), and that is trivial to compute. 
For the second challenge, we will have to limit the number of points on the ball in order to be able to compute $\gamma$ in a reasonable amount of time. Thus we may take a random sampling of the hypercube as a necessary approximation. Finally, if the output is not just a scalar, for which case interpreting $\gamma$ is natural,  but rather multidimensional, one must decide whether some additional transformation is needed for interpretation. To take a weather example, if the model output is a 3-dimensional wind velocity, then   $\overrightarrow{\gamma}$ represents the instability in velocity, and one might want to take its magnitude or angle with respect to north to interpret as speed instability or directional bias, respectively.

For the purpose of demonstration, we choose here to focus on high-dimensional image-classification models, due to popular practicality and ease of interpretation. The inputs (pixels) and outputs (class logits) are typically both high dimensional, and would thus serve to illustrate the behavior of any other high dimensional model as well. In particular, we consider ResNet-50 \cite{resnet50} and the Vision Transformer \cite{vit}. 

\subsection{ResNet-50 and ViT}

ResNet-50 and Vision Transformer (ViT) are image classifiers trained on 1000 distinct classes. To keep things manageable in this short work, we employ the following restrictions:
\begin{itemize}
\item Data is restricted to grayscale images: the value of every pixel is thus an integer from 0 to 255
\item Images are rescaled to 100x100 resolution: each image will thus be a 10000-dimensional vector
\item Approximate simplices: 10000-simplices are well-approximated as 1-hot vectors on the 10000-dim unit-hypercube as noted above; we scale each vector to magnitude 100 which amounts to a significant tone change at the position of the corresponding pixel. To increase ball coverage, we will use the simplices together with their reflections (anti-simplices).
\item Random sampling from simplices: rather than compute at all 20000 ball points (10000 simplex + 10000 anti-simplex) for each image, we take a random 0.1\% sampling of such. 
\item We compute $\gamma$ only in the logit occupying the dimension of its predicted class label 
\end{itemize}
These approximations are done for speed of experimentation and may seem very limiting, but the results below justify them.

We apply these models to an animals test set \cite{animals}, consisting of over 20k color images of animals in various resolutions from a pre-determined set of 10 classes (dog, horse, elephant, butterfly, chicken, cat, cow, pig, spider, squirrel). Then, as described above, we rescale each image to 100x100 pixels and convert to grayscale before computing $\gamma$ in its predicted class logit dimension. We evaluate 100 images per class, which will be sufficient to see the trends in robustness.

For each image, we also execute an adversarial search process wherein we follow the gradient of $\gamma$ for 25 iterations, recording the final image and its predicted class (see Algorithm~2 below). Each such image being only 25 pixels disparate from its original form, a change in class label is interpreted as "instability" in the original image, reminiscent of earlier gradient-based adversarial work \cite{wu2021beating}. Note this procedure is actually stochastic gradient ascent as our $\gamma$-computation is based on random sampling of the hypercube.

\begin{algorithm}
\caption{$\gamma$-Stochastic Adversarial Search at a point x in feature space}
\begin{algorithmic}[1]
\Procedure{AdversarialSearch($x$,$r$,$N$)}{}
\State $\textrm{currPoint} \gets x$
\State $\textrm{numSteps} \gets N$
\For{\texttt{each $\mathrm{step}$ in $\mathrm{numSteps}$}}
\State currGammas $\gets$ \{\}
\State $\textrm{ballPoints} \gets \mathrm{Ball}(\mathrm{currPoint},r)$
\For{\texttt{each $\mathrm{point}$ in $\textrm{ballPoints}$}}
\State $\textrm{currGammas[point]}$ $\gets$ $\gamma(\mathrm{point},r)$
\EndFor
\State $\textrm{currPoint} \gets argmax(\textrm{currGammas})$
\EndFor
\State \Return $\textrm{currPoint}$
\EndProcedure
\end{algorithmic}
\label{alg:stochastic}
\end{algorithm}

\begin{figure}
\centering
\fbox{\rule[-.5cm]{0cm}{0cm} 
\includegraphics[width=0.5\linewidth]{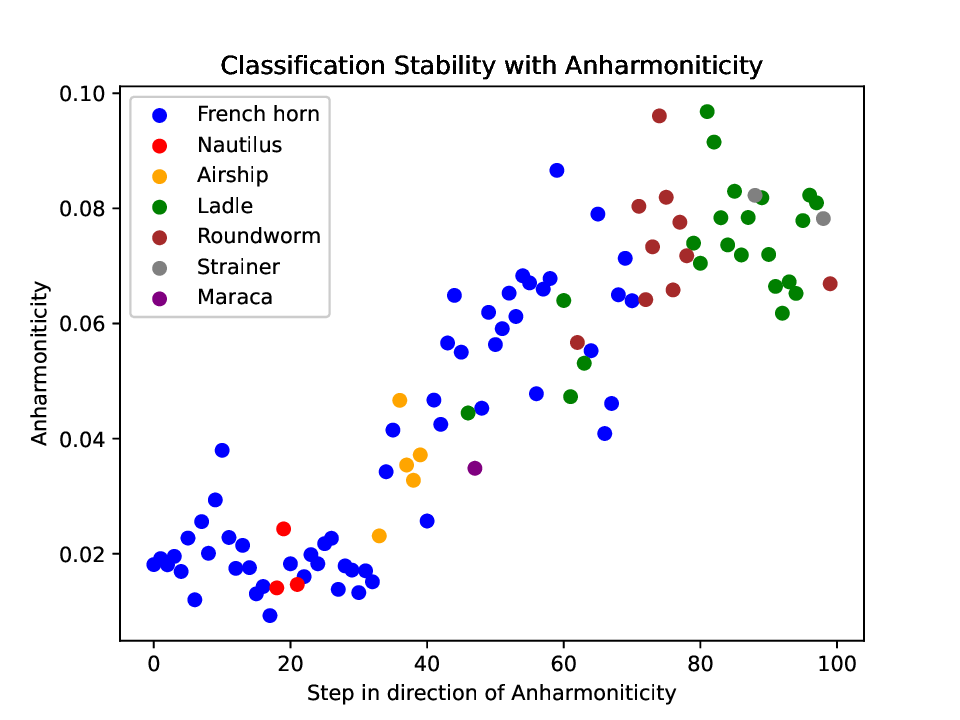}
\includegraphics[width=0.5\linewidth]{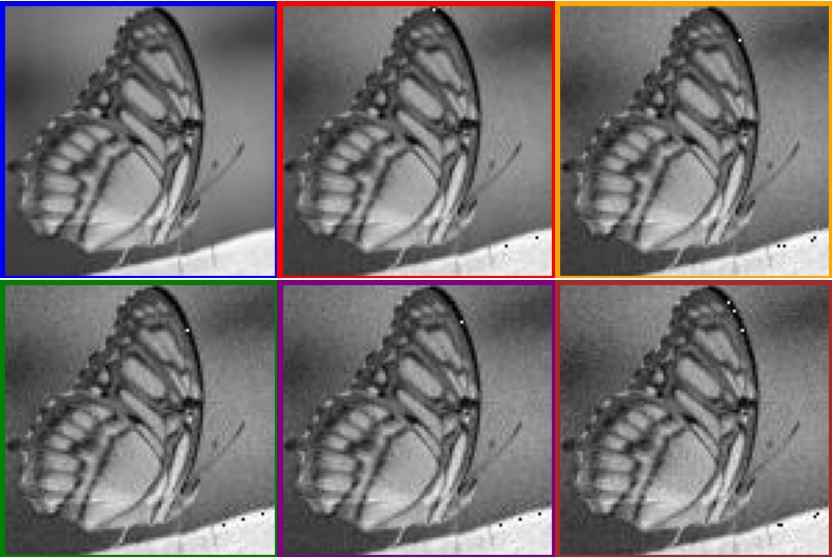}
\rule[-.5cm]{0cm}{0cm}}
  \caption{Demonstration of adversarial search procedure: following the stochastically increasing gradient of anharmoniticity brings out classification instability. The image at step N differs from the original image by N pixels.}
 \label{fig:butterfly}
\end{figure}

We chose 25 as the number of steps to execute as preliminary experiments showed this is generally the number of pixels one needs to change for the data and models under review before adversarial examples appear. Figure~\ref{fig:butterfly} shows one of such experiments where we execute the adversarial search for 100 steps, the original class label changing ever more frequently along that path of (stochastically) increasing $\gamma$. This procedure is actually very effective for quickly and reliably finding adversarial attacks on any input image.

\begin{figure}[ht]
\centering
\subfloat{\begin{overpic}[width=0.19\textwidth]{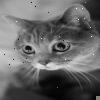}\put
(5,5) {\color{white}wash basin}\end{overpic}}\hfill
\subfloat{\begin{overpic}[width=0.19\textwidth]{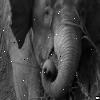}\put (5,5) {\color{white}roundworm}\end{overpic}}\hfill
\subfloat{\begin{overpic}[width=0.19\textwidth]{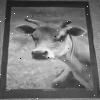}\put (5,5) {\color{white}pig}\end{overpic}}\hfill
\subfloat{\begin{overpic}[width=0.19\textwidth]{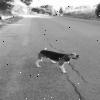}\put (5,5) {\color{white}tandem bicycle}\end{overpic}}\hfill
\subfloat{\begin{overpic}[width=0.19\textwidth]{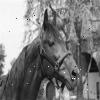}\put (5,5) {\color{white}muzzle}\end{overpic}}\hfill\\
\vspace{1mm}
\subfloat{\begin{overpic}[width=0.19\textwidth]{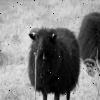}\put (5,5) {\color{white}marmot}\end{overpic}}\hfill
\subfloat{\begin{overpic}[width=0.19\textwidth]{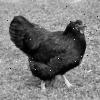}\put (5,5) {\color{white}gorilla}\end{overpic}}\hfill
\subfloat{\begin{overpic}[width=0.19\textwidth]{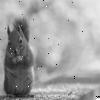}\put (5,5) {\color{white}mousetrap}\end{overpic}}\hfill
\subfloat{\begin{overpic}[width=0.19\textwidth]{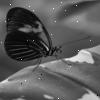}\put (5,5) {\color{white}radiator grille}\end{overpic}}\hfill
\subfloat{\begin{overpic}[width=0.19\textwidth]{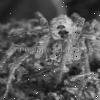}\put (5,5) {\color{white}roundworm}\end{overpic}}\hfill
\caption{Examples of adversarial examples in ResNet-50 from following gradient of $\gamma$ for 25 steps. Each image was originally correctly classified, but changed classes with modification of 25 scattered pixels as shown.}
\label{fig:adversarials}
\end{figure}

Table~\ref{tab:logits} shows the resulting statistics of applying ResNet-50 and ViT to the data.
Looking at the Accuracy and Stability columns, one notices that ViT is generally more accurate and robust than ResNet-50, but not for all classes: e.g., for Cow (ResNet is more robust) and Squirrel (ResNet is more accurate and robust). 
See Figure~\ref{fig:adversarials} where we show examples from each class where the predicted class radically changes after changing just 25 scattered pixels according to our adversarial scheme. Presumably these models behave more stably on larger color images, but it is useful to see how they behave on out-of-domain data, and $\gamma$ gives you a way to measure that.

\begin{table}
\centering
\caption{Average Accuracy and Stability metrics of 10 animal classes (100 images each) for ResNet-50 and Vision Transformer (ViT). ${\overline {\cal P}_C}$ is average softmax probability for the class, and Stability measures percentage of samples with stable classification after N=25 adversarial steps. ViT is typically more accurate and robust but not always (e.g. Cow and Squirrel classes).}
\label{tab:logits}
\begin{tabular}{|l|r|r|c|c|c|c|c|c|} 
\hline
 Class & Model &  $\overline{L}_{C}$ & $\overline{L}$ & ${\overline {\cal P}_C}$ & $\overline{\gamma}$ &  $ {\overline {\cal P}_C} e^{-N\overline{\gamma}}  $  & Accuracy \% & Stability \%\\ 
\hline
Chicken  & ResNet  & -0.12 & -9.35 & 0.911 & 0.042 & 0.32     & 34 & 57 \\
  & ViT & 8.91 & -6.9 $\cdot 10^{-5}$  & 0.881 & 0.027 & 0.45      & \textbf{70} & \textbf{68} \\
\hline
Butterfly  & ResNet  & 0.038 & -9.09 & 0.929 &   0.038 & 0.36    & 36 & 52 \\
  & ViT & 8.84 & -3.0 $\cdot 10^{-5}$  & 0.873 & 0.034 &  0.37     & \textbf{60} & \textbf{67} \\
\hline
Sheep  & ResNet  & 0.05 & -9.86 & 0.953 & 0.037 & 0.38     & 40  & 58 \\
  & ViT &  8.07 & 4.5 $\cdot 10^{-5}$   & 0.762 & 0.022 & 0.44       & \textbf{71} & \textbf{81} \\
\hline
Cat  & ResNet  & 0.96 & -9.55 & 0.973 & 0.054 &  0.25    & 80 & 73 \\
  & ViT & 10.18 & 2.8 $\cdot 10^{-4}$  & 0.963 & 0.082 & 0.12      & \textbf{83} & \textbf{76} \\
\hline
Dog  & ResNet  & 0.99 & -10.03 & 0.984 & 0.040  & 0.36    & 87 & 72 \\
  & ViT & 11.07 & 1.2 $\cdot 10^{-4}$ & 0.985 & 0.039  & 0.37     &  \bf{94} & \bf{80} \\
\hline
Elephant  & ResNet  & 0.96 & -10.08 & 0.984 & 0.041 & 0.35     & 89 & 75 \\
  & ViT & 12.98  & 1.8 $\cdot 10^{-6}$& 0.995 & 0.027 & 0.51     & \textbf{90} & \textbf{81} \\
\hline
Horse  & ResNet  & 1.71 & -9.56 & 0.987 & 0.038 & 0.38     & 67 & 72 \\
  & ViT & 8.95 & 3.9 $\cdot 10^{-5}$ & 0.885  & 0.020 & 0.54      & \textbf{89} & \textbf{88}  \\
\hline
Spider  & ResNet  & 1.94 & -9.74 & 0.992 & 0.035 & 0.41      & 66 & 78 \\
  & ViT & 10.99 & 2.6  $\cdot 10^{-5}$ & 0.983  & 0.029 & 0.48      & \textbf{73} & \textbf{82} \\
\hline
Cow  & ResNet  & 1.84 & -10.00 & 0.993 & 0.033 & 0.44     & 79 & \textbf{92} \\
  & ViT & 9.72 & -2.2 $\cdot 10^{-5}$  & 0.944 & 0.022 & 0.54     &  \textbf{90} & 90 \\
\hline
Squirrel  & ResNet  & 4.08 & -9.72 & 0.999 & 0.044 & 0.33       & \textbf{79} & \textbf{88} \\
  & ViT & 11.72 & 1.0 $\cdot 10^{-4}$ & 0.992 & 0.044 & 0.33     & 77 & 84 \\
\hline
\end{tabular}
\end{table}

 This goes to show that robustness is not something which is usually constant across data space, nor necessarily correlated with accuracy. Rather, it is the class softmax probability ${\cal P}_C$, controlled by the class logit ($L_C$) and average logit ($\overline{L}$), 
\begin{equation}
Prob (softmax)_C \equiv {\cal P}_C \equiv \frac{e^{L_C}}{\Sigma_{i=1..N_{classes}} e^L_i} \approx \frac{e^{L_C}}{e^{L_C} + (N_{classes}-1) \cdot e^{\overline{L}}}
\end{equation}
which, after a certain number N of gradient steps, reduces the class logit on average, from $L_C$ to $L_C-N\gamma$  so the adjusted probability becomes
\begin{equation}
{\cal P'}_C  \approx \frac{e^{L_C-N\gamma}}{e^{L_C-N\gamma} + (N_{classes}-1) \cdot e^{\overline{L}}} \approx {\cal P}_C e^{-N\gamma} 
\end{equation}

In aggregate, we can already see in Table~\ref{tab:logits} that the average value of this quantity, ${\overline {\cal P}_C} e^{-N\overline{\gamma}}  $, shows a reasonable correlation with class stability, but exceptions do arise because the stochastic nature of the gradient ascent procedure does not guarantee that the class logit changes by $N\gamma$ after N steps\footnote{The Cat class, for example, has average logit changes of 0.67 and 0.81 for ResNet and ViT, respectively, while $N\overline{\gamma}$ is overestimating at 1.34 and 2.05, respectively. On average across classes it is close: the average logit drift after $N=25$ iterations is $0.88(2)$ while N times the average $\gamma$ is $0.94$. }.

\begin{figure}
\centering
\fbox{\rule[-.5cm]{0cm}{0cm} 
\subfloat[]{\includegraphics[width=0.4\linewidth]{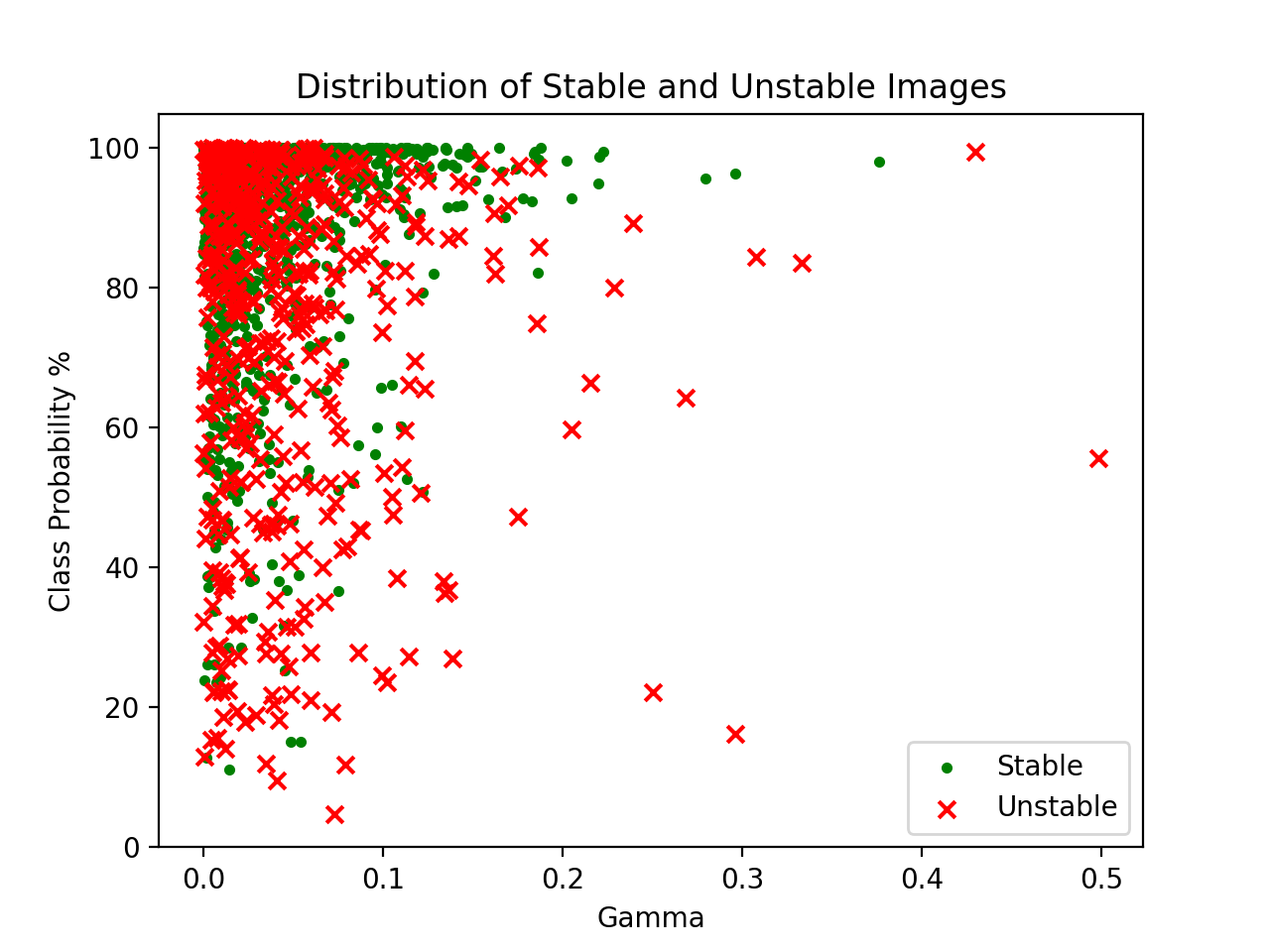}}
\subfloat[]{\includegraphics[width=0.4\linewidth]{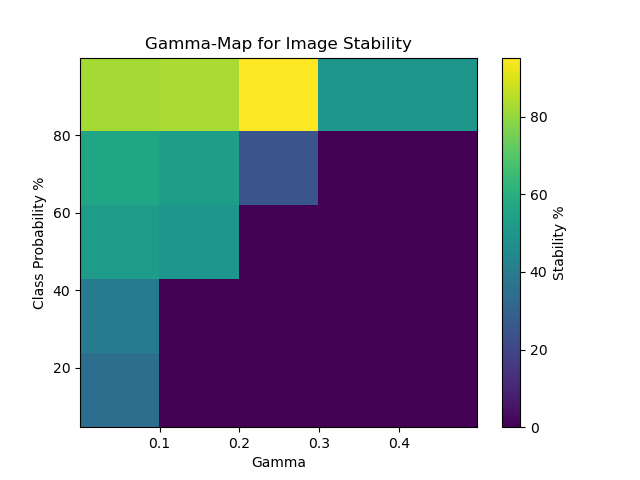}}
\rule[-.5cm]{0cm}{0cm}}
  \caption{(a) Plotting predicted class probability $\cal P$ and $\gamma$ for 1000+1000 images classified with ResNet and ViT shows the unstable images tend to dominate high-$\gamma$/low-$\cal P$ regions. (b) Correlation of image stability (\% images that are stable)  with $\cal P$ and $\gamma$ as a density plot serves as a "Gamma Map" for this application. }
 \label{fig:resnet-corr}
\end{figure}

The way to use this metric in real time inference, of course, would be on an image-by-image basis, where $\cal P$ and $\gamma$ are outputs of the model along with the predicted class label. 
On that note, plotting the measured values of Class Probability and $\gamma$ for each image, as well as whether it is stable or not after 25 iterations, we obtain a ``Gamma Map" (Figure~\ref{fig:resnet-corr}). As a practical tool, this plot allows one to immediately gauge whether a predicted classification is likely to be stable just from measuring its Class Probability $\cal P$ and $\gamma$, i.e., without having to do a full adversarial search.

\section{Discussion}

We hope the reader sees that the theory and computation of anharmoniticity ($\gamma$) is quite simple, straightforward, and applicable to any model function. The computational complexity is quite low, requiring only standard vector manipulations, so should be relatively fast to compute. Since the computation of $\gamma$ does require running several extra model predictions per point, it may lag real-time events slightly  depending on the model latency, but will probably still be ``realtime enough" to provide fast feedback to a monitoring system.

There is algorithmic freedom in choosing the ball approximation and radius: we argue for a simplex ball as it is a closed-form fast computation in virtually any number of dimensions, but we cannot rule out a better choice. Higher dimensional models, increasingly common in the ML community as computational power advances, are at any rate served well by a hypercube which incurs no overhead. The radius $r$ should be small enough to count as ``local", but not so small as to submerge  meaningful variations beneath feature space noise. Generally, one chooses $r$ based upon what level of such noise is expected, and each model+feature space setup will thus probably demand some fixed, reasonable choice; judging, however, by the variation in $\gamma$ for various $r$ in Fig.~\ref{fig:radii}, the exact choice is not critical.

One can use Eqn.~\ref{gamma} as-is for $\gamma$, as we did for the low-dimension examples with scalar outputs, or use a projection of multidimensional $\gamma$ as we did for the image classification, focusing on the classification logit dimension only. In general, one could use some other transformation or reduction of the full $\gamma$, again depending on the model and application. If additional degrees of freedom appear in this procedure, then one may generically obtain some kind of multidimensional "Gamma Map" mapping these variables to a scalar stability metric, as we saw for example in our treatment of image classification models with the Gamma Map shown in Fig.~\ref{fig:resnet-corr} --- this allows immediate point-wise identification of robustness.  

Turning to using $\gamma$ in practice, for the training examples we showed $\gamma$ could actually be used on the validation set to provide additional feedback about overfitting, balancing  $\gamma$, validation loss, and training loss. It can also rank models trained on the same data.  While for the inference examples (with ResNet and ViT), one might include $\gamma$ in a monitoring system as live data comes in (or batched daily) to provide update on data drift and the need to refit in certain areas of feature space.

As a metric to publish with a model's quality card, one can envision reporting $\gamma$ for different data sets, indicating where a model is expected to more perform strongly against adversarial attacks.
We saw that following the path of anharmoniticity from a base image can efficiently provide an adversarial example, so the fewest number of steps on this path to an adversarial example can be used to complement adversarial distance, another metric of stability.

Finally as a standard for ML model quality, functional standards are translatable, shareable, and optimizable across the industry; they may even point to certain mathematical truths pertaining to optimal ML systems.

\section{Conclusions}

In this brief work we introduced the method of Harmonic Robustness, computing anharmoniticity ($\gamma$) as a measure of a model's robustness. It is computationally simple, fast, intuitive (explainable), and effective even in the face of approximate application (e.g. sampling over the ball). We saw in our image classification examples that $\gamma$ is useful for estimating adversarial weakness of data without actually having to find the adversarial points --- this can save much work for robustness analysis and may be the first such estimation technique of its kind.

Monitoring systems and testing procedures can easily integrate computations of $\gamma$ as an alerting and regression test mechanism, respectively; while there is no reason why model builders should hold back from computing $\gamma$ alongside usual validation loss and other metrics to control overfitting. As a proxy for explainability, this may lead into incorporating other metrics for Responsible AI into the model life-cycle.

We close with emphasizing a possibly trailblazing facet of our work: that one may apply a functional mathematical standard, i.e., conformity to the properties of harmonic functions, to a ML system as a way of assessing its quality and propriety for public usage. Understandably any such standard is limiting, but need not be extreme.  The harmonic standard is indeed limiting and not necessarily ideal, most real-world ML functions being far from harmonic, but we posit that it is better to reference a standard than having no such standard at all, giving AI systems free reign in their inner complexity while relying on conventional external metrics like precision and recall for quality control. For, assuming the white- or gray-box environment is not always going to be available to us,  if we do not devise multiple ways to check models' inner complexity in a black-box environment, we will be giving up too much control over what these systems may surprise us with.

\section*{Acknowledgments}
We would like to thank Sam Hamilton, Hao Yang, and Joydeep Mitra for early discussion of our technique; we thank Yiwei Cai for critical comments to improve our mathematical treatment.

\bibliography{harmonic_robustness_visa}
\bibliographystyle{unsrt}

\appendix

\section{Gamma correlates with Decision Boundary Length}
\label{app:boundary}
\begin{figure}
\centering
\fbox{\rule[-.5cm]{0cm}{0cm} 
\includegraphics[width=1\linewidth]{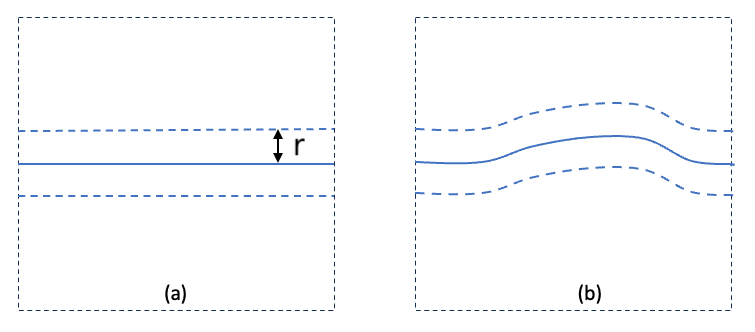}
\rule[-.5cm]{0cm}{0cm}}
  \caption{Two areas of decision space, one with a short boundary (a) and a longer boundary (b). For a given radius $r$, anharmoniticity is non-zero between the dotted lines a distance $r$ from the boundaries. }
 \label{fig:region3}
\end{figure}

In Section~\ref{sec:ml} we stated that $\gamma$ indicated the 'wiggliness' of the decision boundary. Let us argue this more mathematically here, for the case of a smooth decision boundary in feature space, that the average value of $\gamma$ is proportional to the "length" of the decision boundary. So for the two areas of decision space shown in Fig~\ref{fig:region3}, we are setting out to show the region on the left (a) should have a higher average anharmoniticity than the one on the right (b), for any given choice of the radius parameter $r$. Note anharmoniticity is nonzero only for points within $r$ of the boundary, as the ball around any point will only have disparate values when part of it crosses the boundary.

Now for a mathematical 'proof' of this. At any given point $x$ in feature space, the anharmoniticity is defined as
\begin{equation}
\gamma(x,r) \equiv \vert f(x) - \frac{1}{V r^n} \int_{B(x,r)} f dV \vert
\end{equation}
for some choice of ball radius $r$.
In any bounded region $R$ of feature space, then, the average anharmoniticity in this region is defined as
\begin{equation}
\label{averagegamma}
\overline{\gamma} \equiv \frac{1}{V_R}\int_R \gamma(x)~ dx
\end{equation}

Note there is no need to  make any assumptions about how the function behaves in feature space, e.g., whether it is smooth or discontinuous, full of pockets of minima and maxima and the like. We simply apply Eqn~\ref{averagegamma} as-is, as a statement of how close the function is to being harmonic, on average.

\begin{figure}
\centering
\fbox{\rule[-.5cm]{0cm}{0cm} 
\subfloat[]{\includegraphics[width=0.4\linewidth]{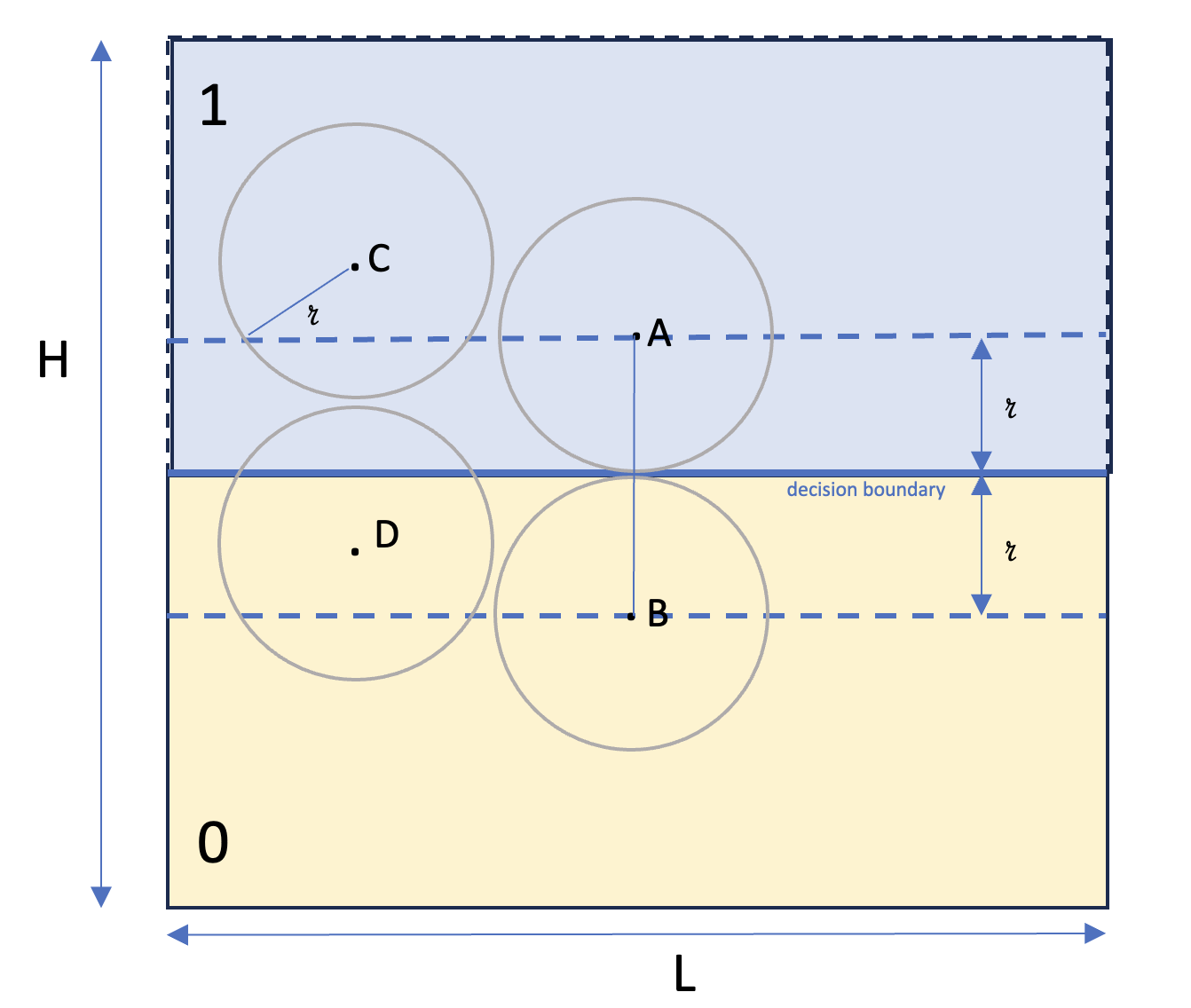}}
\subfloat[]{\includegraphics[width=0.4\linewidth]{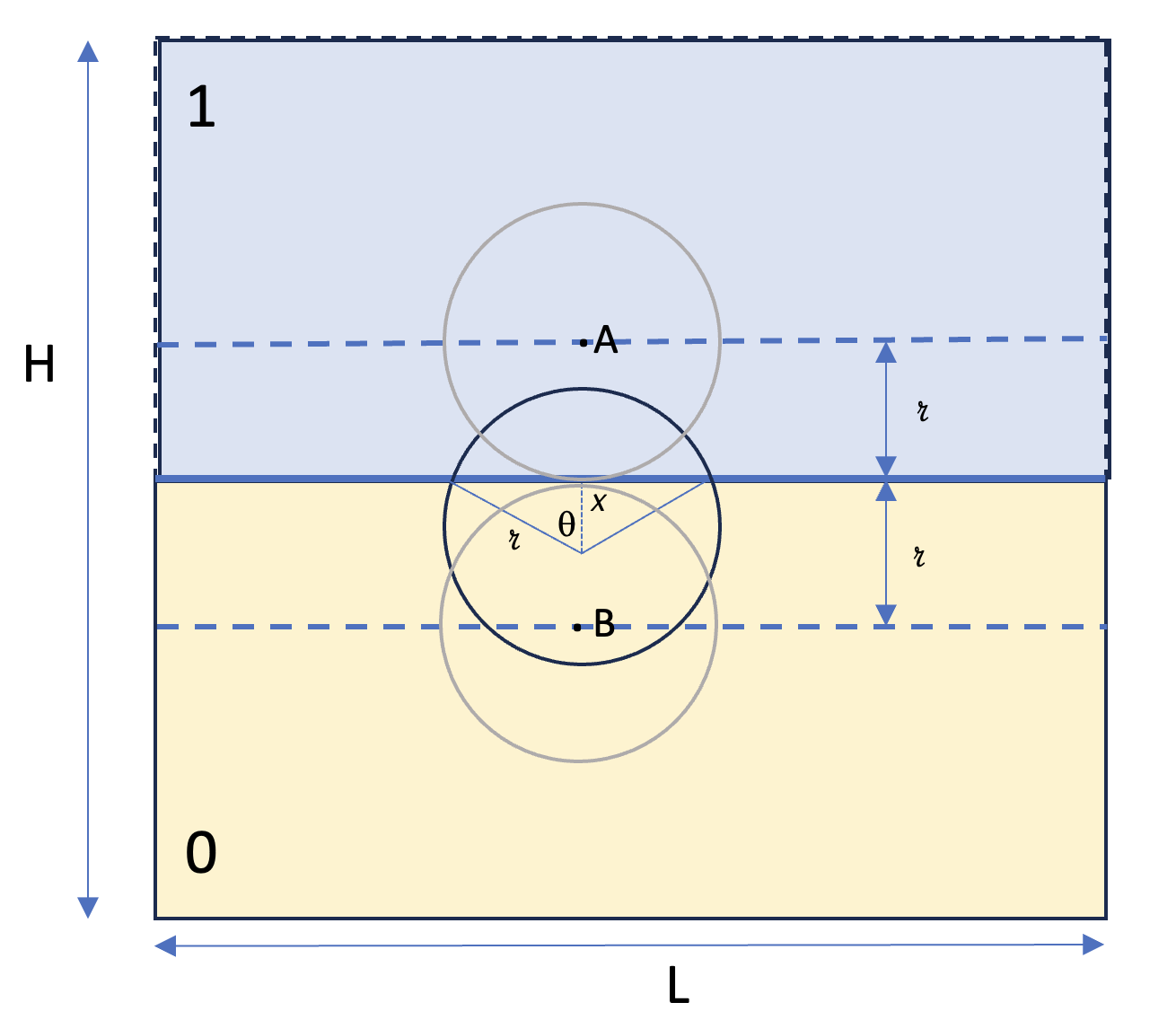}}
\rule[-.5cm]{0cm}{0cm}}
  \caption{Region of 2d feature space of length L and Height H, with sharp decision boundary in the middle. For a given choice of radius $r$, anharmoniticity is nonzero only for points within $r$ of the boundary. }
 \label{fig:region1}
\end{figure}

But for definiteness, consider the case of a binary classifier, with a region where a sharp decision boundary divides feature space into the '1' class and the '0' class (see Fig~\ref{fig:region1}).

At some points in this region, such as point C in Fig~\ref{fig:region1}(a), the anharmoniticity is zero because the ball around point C is completely on one side of the decision boundary. At other points, e.g., point D, the ball crosses the boundary and thus $\gamma$ will be nonzero. Indeed for this local patch of feature space, the only points with non-zero $\gamma$ are those in the locus of points distance $r$ or less from the decision boundary.  Integrating $\gamma$ along the line segment $\overline{AB}$, for example, multiplying by the boundary length $L$, then dividing by the volume of the whole region ($H\cdot L$), gives us the average $\gamma$ in this region.

This integral along $\overline{AB}$ is straightforward: following Fig~\ref{fig:region1}(b), we first integrate a sliding circle from distance $x$ below the decision boundary (0 to $r$), and then again above the decision boundary (0 to $r$); in each case the integrand being the difference between the function at the center of the circle and the average value on the circle:
\begin{eqnarray}
\overline{\gamma} & = & \frac{1}{HL}\left( \vert 0 - \int^r_0 \frac{2~ arccos\frac{x}{r}}{2 \pi}  dx \vert + \vert 1 - \int^r_0 \frac{2\pi - 2~arccos\frac{x}{r}}{2\pi}dx \vert \right) \\
& = & \frac{2}{\pi} \left( x~arccos\frac{x}{r} - r \sqrt{1-(\frac{x}{r})^2}     
\right)^r_0 \\
& = & \frac{2r}{\pi}
\end{eqnarray}

Thus the average $\gamma$ of the straight line region Fig.~\ref{fig:region3}(a) is $L\cdot\frac{2r}{\pi}/(LH)$ = $\frac{2r}{\pi H}$. For the curved line region in Fig.~\ref{fig:region3}(b), the computation along a corresponding path $\overline{AB}$ follows exactly the same except there is \emph{more} decision boundary $L'$ to integrate along ($L' > L$), so overall that region will have greater average $\gamma$ = $L'\cdot\frac{2r}{\pi}/(LH) > \frac{2r}{\pi H}$. QED.

\section{Simplex Computation}
\label{app:simplex}
To compute the n-dimensional simplex about any n-dimensional point $\overrightarrow{p}$, we first compute it about the n-dimensional origin.
One way to do this is to add an auxiliary (n+1)-st dimension and construct the vectors corresponding to the vertices of the (n+1)-dimensional hypercube:
\begin{eqnarray}
v_1 &=& (1,0,0,..0)_{n+1} \\
v_2 &=& (0,1,0,..0)_{n+1} \\
...&=& ...\\
v_{n+1} &=& (0,0,0,..1)_{n+1} \\
\end{eqnarray}

Centering these vectors around the origin by translating by the group average, $\overline{v} = \frac{1}{n+1}(1,1,1,...)_{n+1}$,
\begin{eqnarray}
v'_1 &=& (1-\frac{1}{n+1},\frac{-1}{n+1},..\frac{-1}{n+1})_{n+1} \\
v'_2 &=& (\frac{-1}{n+1},1-\frac{1}{n+1},..\frac{-1}{n+1})_{n+1} \\
... \\
v'_{n+1} &=& (\frac{-1}{n+1},...,\frac{-1}{n+1},1-\frac{1}{n+1})_{n+1} \\
\end{eqnarray}
we then 'rotate away' the auxiliary (n+1)-st dimension by the angle $\theta$  formed by the (n+1)-st unit vector $\hat{n}_1 = (0,0,...,0,1)_{n+1}$  and the normal to the hyperplane formed by the n+1 vectors, $\hat{n}_2 = \frac{1}{\sqrt{n+1}} (1,1,...,1)_{n+1}$ using a generalization of Rodrigues' formula for the rotation by $\theta$  in a hyperplane formed by any two orthonormal vectors $\textbf{n}_1$ and $\textbf{n}_2$:
\begin{equation}
{\cal{R}} = I + (\textbf{n}_2 \textbf{n}_1^T - \textbf{n}_1 \textbf{n}_2^T) sin\theta + (\textbf{n}_1 \textbf{n}_1^T + \textbf{n}_2 \textbf{n}_2^T) (cos\theta - 1)
\end{equation}
with $\textbf{n}_1 = \hat{n}_1$, $\textbf{n}_2 = |\hat{n}_2 - (\hat{n}_2 \cdot \hat{n}_1)\hat{n}_1| = \frac{1}{\sqrt{n}}(1,1,...,1,0)_{n+1}$, and $cos\theta = 1/\sqrt{n+1}$. Thus
\begin{eqnarray}
v''_1 & = & {\cal{R}} v'_1 \\
v''_2 & = & {\cal{R}} v'_2 \\
... & = & ... \\
v''_{n+1} & = & {\cal{R}} v'_{n+1} = (0,0,...,0)_{n+1} \\
\end{eqnarray}
where the last vector is 'rotated away' by construction and can be dropped. The other n vectors form the vertices of the origin-centered, symmetric n-simplex in n-dimensions. Now to form the simplex ball about $\overrightarrow{p}$ as in Algorithm 1, one simply adds them vectorially, scaled to magnitude $r$, i.e., $\overrightarrow{p} + r\textbf{v}''_1$, $\overrightarrow{p} + r\textbf{v}''_2$, etc. The simplex + anti-simplex ball adds the negative displacements in as well, i.e., $\overrightarrow{p} - r\textbf{v}''_1$, $\overrightarrow{p} - r\textbf{v}''_2$, etc.

Now taking the limit of $n \to \infty$, we see that ${\cal R} \to I$, $v' \to v$, and thus $ v'' = {\cal R} v' \to v$, i.e., the n-simplex vertices for high dimensions converge to the vertices of the n-dimensional hypercube. 

\section{Application to Simple Functions}
\label{app:functions}
Let us compare how our harmoniticity metric performs on simple known functions, using two different approximations to the n-ball: one with n-dimensional random vectors, and one using the vertices of the n-simplex.

\begin{figure}
\centering
\fbox{\rule[-.5cm]{0cm}{0cm} 
\subfloat[]{\includegraphics[width=0.5\linewidth]{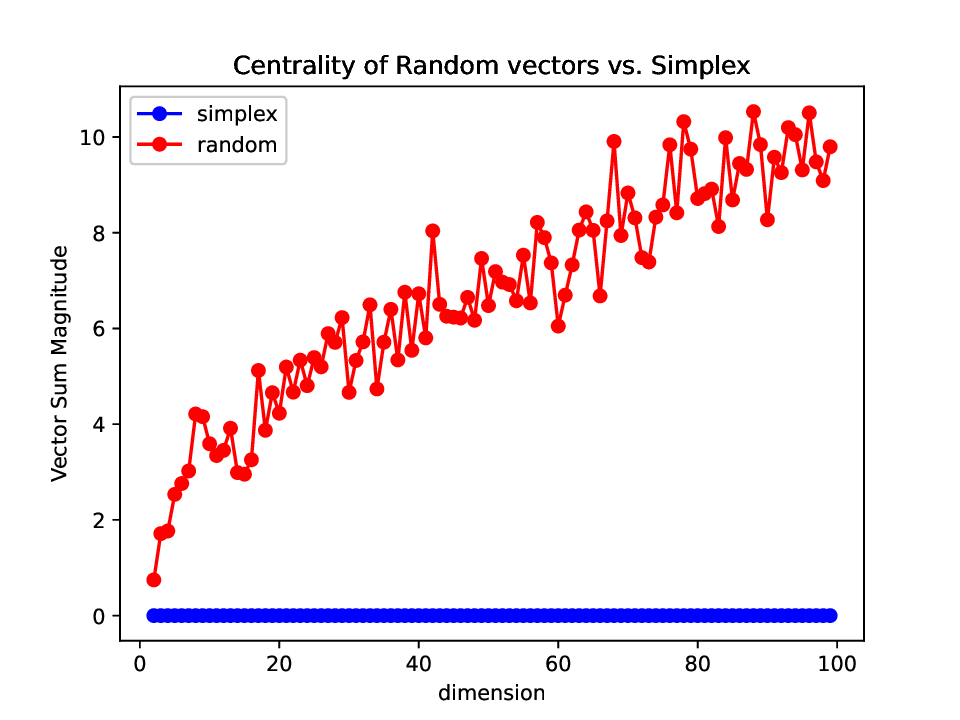}}
\subfloat[]{\includegraphics[width=0.5\linewidth]{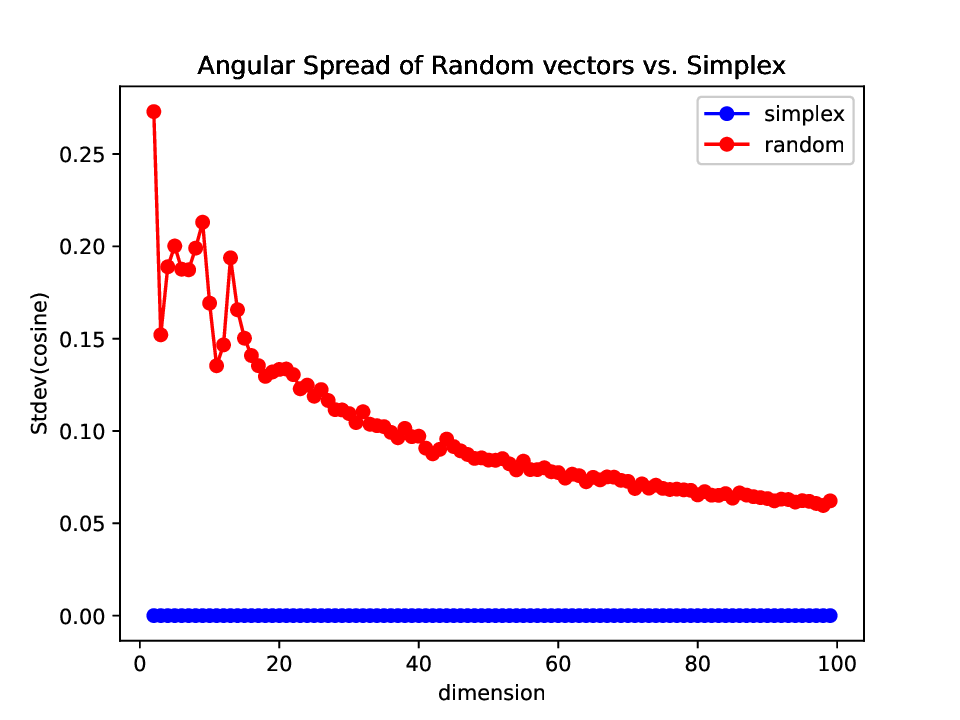}}
\rule[-.5cm]{0cm}{0cm}}
  \caption{Comparison of space coverage between n random vectors and the n vertex vectors of a (n-1) simplex. Random vectors trace out a random walk and actually never achieve symmetry. The simplex points are all consistent with zero, meaning they are exactly centered and spread evenly.}
 \label{fig:angles}
\end{figure}

We first observe how well these two different balls cover space in terms of centrality (how close their vector sum comes to zero) and isotropy (how small is the standard deviation of their angles with respect to a fixed unit vector). For each dimension n, we randomly generate n vectors and compare to the deterministic vectors given by the n-simplex. 
Adding n random vectors of course replicates the random-walk phenomenon, thus centrality is expected to diverge from the origin as $\sqrt{n}$, as we indeed see in Fig.~\ref{fig:angles}.  Isotropy, meanwhile, slowly converges towards zero at high n. For the n-simplices, on the other hand, we have by construction perfect centrality and isotropy.

Now since $\gamma$ is supposed to measure closeness to "harmoniticity", let us demonstrate this actually works on known harmonic and anharmonic functions.
For our first test, consider this pair:
\begin{eqnarray}
f_1(x_0, x_1, ..., x_n) &=& x_0^2 - x_1^2 + x_2^2 - ... - x_n^2 \\
f_2(x_0, x_1, ..., x_n) &=& x_0^2 + x_1^2 + x_2^2 + ... + x_n^2 \\
\end{eqnarray}
The first function $f_1$ is easily seen as harmonic (for even n) since the 2nd derivative of each term is $\pm 2$ which sums to 0, while the second function $f_2$ is similarly not harmonic as all 2nd derivatives are positive. Choosing 1000 points randomly within the n-dimensional unit-hypercube, constructing an approximate  ball around each (either randomly or with simplices as described), we compute the average anharmoniticity (i.e., $\gamma$) as per Algorithm 1.

\begin{figure}
\centering
\fbox{\rule[-.5cm]{0cm}{0cm} 
\subfloat[]{\includegraphics[width=0.5\linewidth]{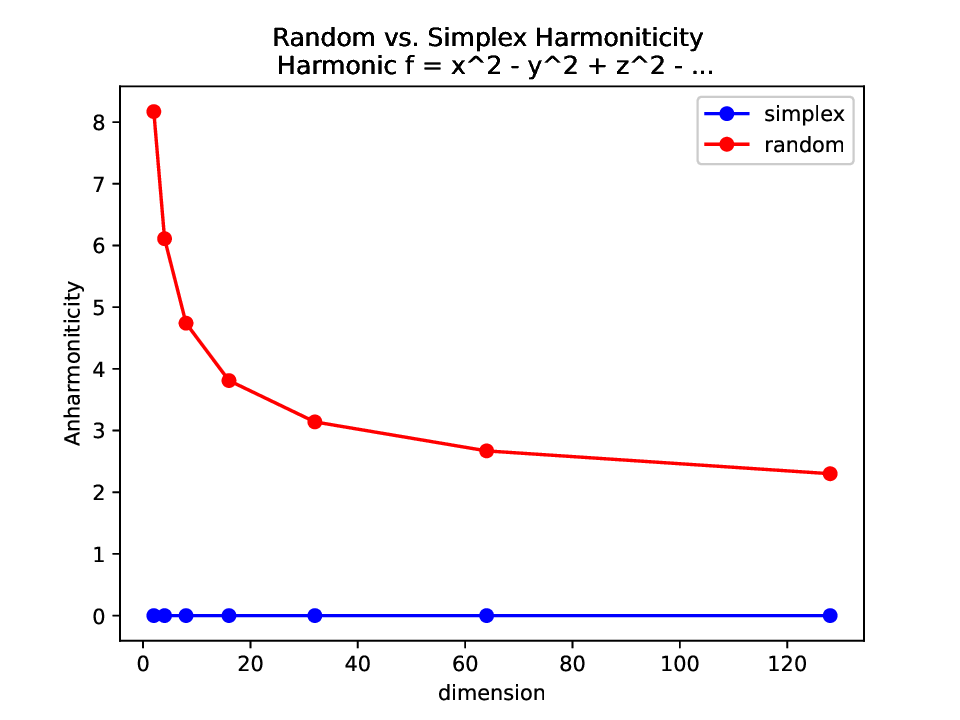}}
\subfloat[]{\includegraphics[width=0.5\linewidth]{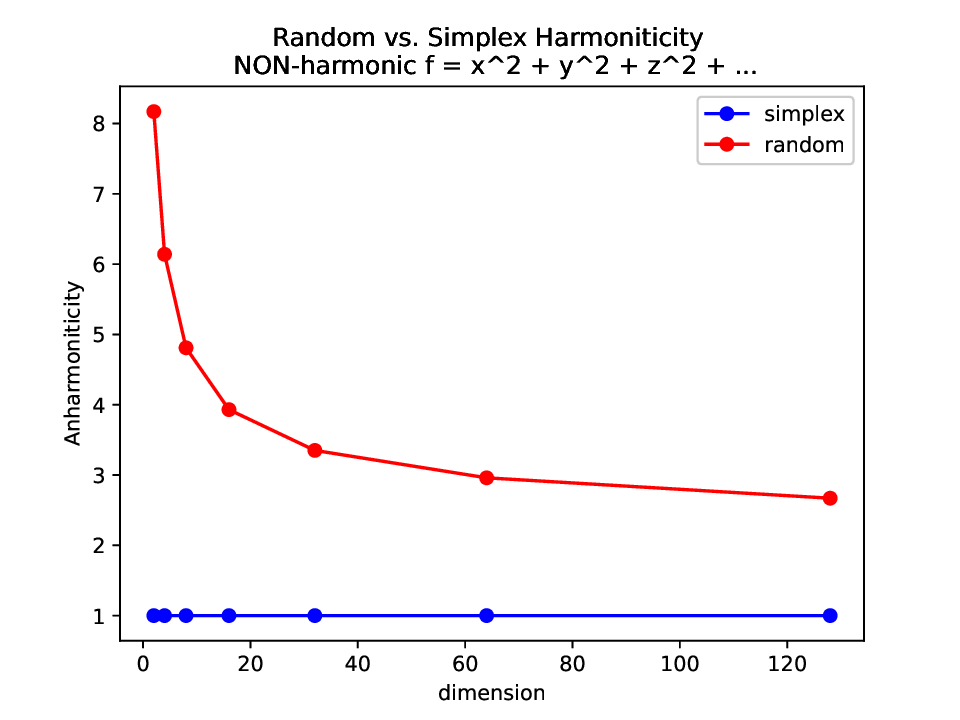}}
\rule[-.5cm]{0cm}{0cm}}
  \caption{Measuring harmoniticity of a known harmonic and anharmonic function, using   n random vectors or the n vertex vectors of a (n-1) simplex. The simplex method happens to work ideally here in all dimensions, while random vectors only very slowly improve at higher dimensions.}
 \label{fig:harmonic}
\end{figure}

As seen in Fig. \ref{fig:harmonic}, the simplex method actually gives the ideal anharmonic value of 0 for the harmonic function, and 1 for the anharmonic function, for all n tested. The random method does not distinguish the functions as well, although this might improve for higher n (or more random vectors).

\begin{figure}
\centering
\fbox{\rule[-.5cm]{0cm}{0cm} 
\includegraphics[width=0.5\linewidth]{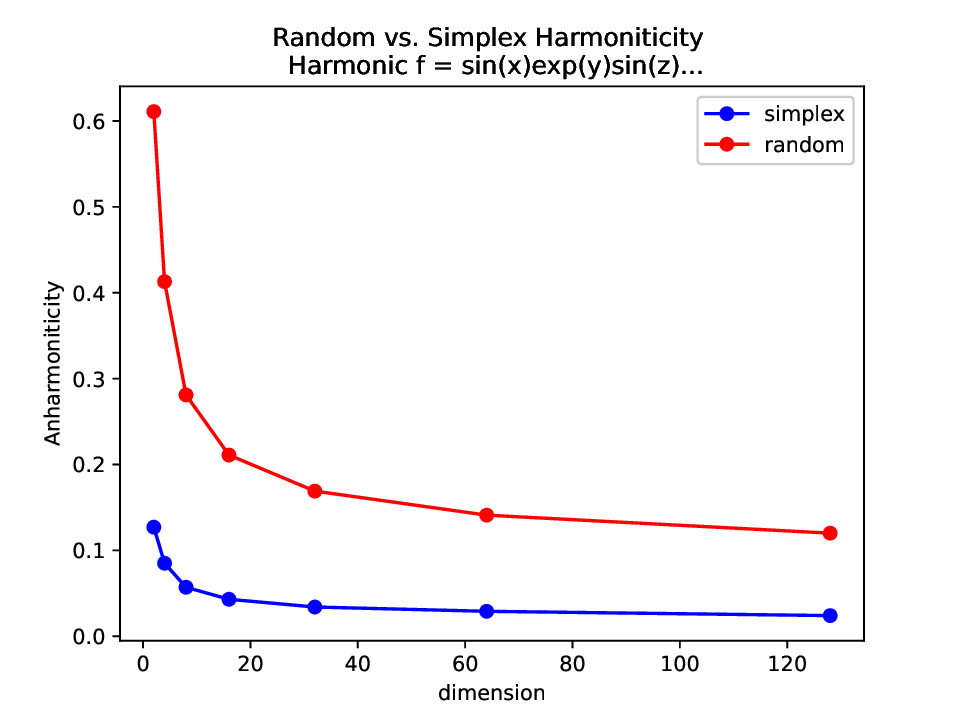}
\includegraphics[width=0.5\linewidth]{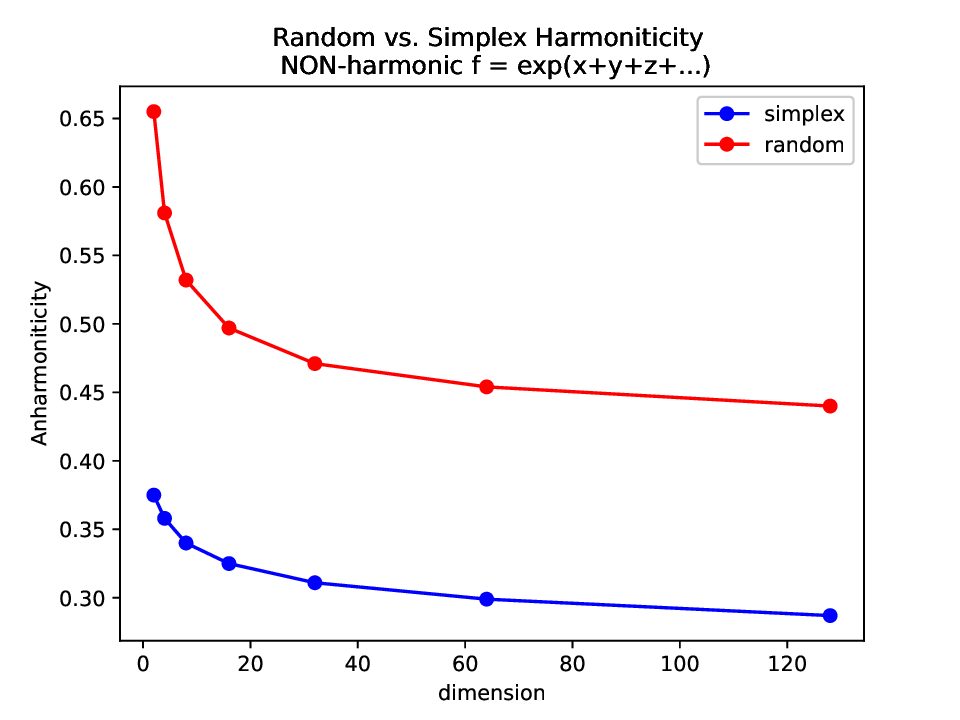}
\rule[-.5cm]{0cm}{0cm}}
  \caption{Measuring harmoniticity of another known harmonic and anharmonic function. The simplex method is still better than random vectors, and shows a significantly lower anharmoniticity for the harmonic function.}
 \label{fig:harmonic2}
\end{figure}

Testing on another harmonic-anharmonic function pair:
\begin{eqnarray}
f_3(x_0, x_1, ..., x_n) &=& sin(x_0) e^{x_1} sin(x_2) e^{x_3} ... e^{x_n} \\
f_4(x_0, x_1, ..., x_n) &=&  e^{x_0 + x_1 + x_2 + ... + x_n}\\
\end{eqnarray}

we repeat the comparison and again the simplex method is superior  in Fig. \ref{fig:harmonic2}. While not getting exactly $\gamma=0$  for $f_3$, it is still significantly lower than that of $f_4$.  

So for these simple cases the technique seems to work on pure functions in any dimension, and especially well with the simplex method.
Actual ML functions that do something interesting (e.g., classifiy MNIST digits, predict credit card fraud, etc.) will of course be much more complicated functions that will have different convergence rates. But based on the above basic observations, we go with the simplex method to most efficiently measure harmoniticity.

\end{document}